\documentclass{3dim-article}

\journal{3dim}
\usepackage{hyperref}

\usepackage{url}

\articletype{Research Article}

\usepackage[normalem]{ulem}

\begin{document}

\title{Accurate Eye Tracking from Dense 3D Surface Reconstructions using Single-Shot Deflectometry}

\author{Jiazhang Wang\authormark{1,2}, Tianfu Wang\authormark{3}, Bingjie Xu\authormark{4}, \\ Oliver Cossairt\authormark{1,3}, and Florian Willomitzer\authormark{1,2,4,*}}

\address{
\authormark{1}Wyant College of Optical Sciences, University of Arizona, Tuscon, AZ, 85721\\
\authormark{2} Department of Electrical and Computer Engineering, Northwestern University, Evanston, IL, 60208\\
\authormark{3}Department of Computer Science, ETH Zürich, Zürich, Switzerland, 8092\\
\authormark{4}Department of Computer Science, Northwestern University, Evanston, IL, 60208\\

}
\email{\authormark{*}fwillomitzer@arizona.edu }

\begin{abstract}
Eye-tracking plays a crucial role in the development of virtual reality devices, neuroscience research, and psychology. 
Despite its significance in numerous applications, achieving an accurate, robust, and fast eye-tracking solution remains a considerable challenge for current state-of-the-art methods. While existing reflection-based techniques (e.g., "glint tracking") are considered \textcolor{black}{to be very} accurate, their performance is limited by their reliance on sparse 3D surface data acquired solely from the cornea surface.

In this paper, we rethink the way how specular reflections can be used for eye tracking: We propose a novel method for accurate and fast evaluation of the gaze direction that exploits teachings from single-shot phase-measuring-deflectometry (PMD). In contrast to state-of-the-art reflection-based methods, our method acquires dense 3D surface information of both cornea and sclera within only one single camera frame (single-shot). \textcolor{black}{For a typical measurement, we acquire  $>3000 \times$ more surface reflection  points (``glints'') than conventional methods. We show the feasibility of our approach with experimentally evaluated gaze errors on a realistic model eye below only $0.12^\circ$. Moreover, we demonstrate quantitative measurements on real human eyes in vivo, reaching accuracy values between only $0.46^\circ$ and $0.97^\circ$. }

\end{abstract}

\section{Introduction}

Eye tracking is an enabling technology\cite{holmqvist2011eye} that is widely involved in different fields of application. In virtual/augmented reality (VR/AR), it enables important functions, such as foveated rendering, compensation of the vergence-accommodation conflict, or interactions with the user interface or virtual avatars\cite{adhanom2023eye}. In clinical applications, it helps with the diagnosis of disorders \cite{clark2019potential}, or facilitates improved surgical training\cite{merali2019eye}. Moreover, eye tracking is frequently used in psychology research \cite{rahal2019understanding} and cognitive neuroscience \cite{hessels2019eye}.

Current state-of-the-art optical eye tracking approaches can be divided into two groups: image-based methods \cite{nishino2004world,lu2011head,lu2011inferring,li2015gaze,valliappan2020accelerating} and reflection-based methods\cite{coutinho2006free,hennessey2009improving,mestre2018robust,liu2022method,beymer2003eye,shih2004novel}. 
Image-based methods detect features such as the pupil, iris, veins, or limbus in 2D  images of the eye and use them to estimate the gaze direction. This can be done, e.g.,  by exploiting geometrical relations in the captured eye image (iris center, pupil eccentricity, etc.)\cite{li2015gaze,lu2011inferring,tan2002appearance,lu2011head}, or via deep learning\cite{valliappan2020accelerating,baluja1993non,krafka2016eye,zhu2017monocular}. The density of 2D image features is, however, relatively sparse for all cases, which potentially limits the quality of the reconstruction. Moreover, the transparent property of the cornea makes it difficult for image-based methods to extract accurate features in these regions (e.g., from the underlying iris) since the refraction of light at the cornea surface is different at different viewing angles\cite{patel2019refractive}.
In contrast, state-of-the-art reflection-based methods directly or indirectly measure the 3D coordinates of the cornea surface \textcolor{black}{via reflection of point lights} and use the respective information to calculate the gaze. `Glint tracking' is the most prominent method in this class \cite{beymer2003eye,coutinho2006free,zhu2007novel,mestre2018robust}: Around 2-12 infrared point light sources are spatially arranged in front of the eye (e.g., in a circle) to be reflected by the cornea surface. Combined with other measured 2D features, such as the pupil center, the gaze direction can be estimated. \textcolor{black}{Reflection-based methods typically reach higher accuracy values than image-based methods. According to \cite{adhanom2023eye, hansen2009eye,kar2017review}, the very best reported errors are in the $ \sim 0.5^\circ -1.5^\circ$ range. However, the authors of \cite{adhanom2023eye} also acknowledge that these numbers are hard to assess, especially for the listed errors of commercially available eye-tracking devices.  In many cases, only the best values under ideal conditions are reported \cite{adhanom2023eye}, the best performance can only be reached at a specific region in the field of view, or the used error metrics are not clearly quantified. All this can make the results of the respective eye trackers hard to reproduce in realistic settings and can hinder their widespread adoption of the respective techniques. Moreover, even a very good value of $1^\circ$ can still leave} room for improvement for some applications. For example, as 
VR/AR headsets might reach the targeted ``retinal resolution'' in the future \cite{zhao2023retinal},  a $ 1^\circ$ error would correspond to $>60$ \textcolor{black}{display pixels}, which could still pose challenges when the navigation through the user interface should happen purely gaze-based with ``mouse precision''. At larger standoff distances of, e.g., $15m$, a $1^\circ$ gaze error leads to a lateral localization error of $ \sim 25cm$, which is already too large, e.g., to exactly determine a focused product in a supermarket shelf or to interact with a virtual avatar in a larger crowd.

\textcolor{black}{This paper is motivated by the question of whether sampling the eye surface at more points (i.e., with more ``glints'') can lead to higher accuracy due to the potentially higher information content in one measurement. This is especially interesting, as current methods do not acquire more than $\sim 12$ sparse surface points from $\sim 12$ glints \cite{mestre2018robust}.} The relatively low number of data points is caused by the conceptual design of the respective glint tracking method: Its susceptibility to ambiguities makes it difficult to increase the number of point sources. \textcolor{black}{Moreover, most methods require the glint reflections to stay solely on the cornea and to be precisely arranged around the pupil, which can additionally restrict their flexibility.}

In this paper, we \textcolor{black}{introduce a novel method for} reflection-based eye tracking \textcolor{black}{that} \textit{radically} increases the number of measured eye surface points from $\sim 12$ to over $40,000$. We pair this significant increase with novel sophisticated algorithms that calculate the gaze direction fast and computationally efficient. Our method to acquire 3D information about the eye surface is based on phase measuring deflectometry (PMD)\cite{knauer2004phase}. PMD was originally invented as an optical metrology method for the dense 3D reconstruction of specular objects, like lenses or astronomical mirrors \cite{knauer2008measuring,hofer2016infrared,faber2012deflectometry,olesch2014deflectometric, hausler2013deflectometry,burke2023deflectometry}.   Modern well-calibrated PMD setups can reach sub-micron \textcolor{black}{shape measurement} accuracy using only inexpensive off-the-shelf equipment (display and camera)\cite{knauer2004phase,faber2012deflectometry,su2012scots,trumper2016instantaneous}. For that reason,  PMD is well known in the metrology community as the big competitor to highly accurate optical interferometry systems.  Recently PMD has been used, e.g.,  in cultural heritage analysis\cite{willomitzer2020hand, li2021low}, or to measure the topography of the human cornea\cite{liang2016single} for medical diagnosis. Its special design for \textcolor{black}{measuring} specular surfaces makes PMD an ideal technique for eye surface measurements which yields several advantages compared to techniques designed to measure matte surfaces, such as fringe projection profilometry \cite{zheng2023fringe}.

In a PMD setup, an off-the-shelf display showing a sinusoidal pattern serves as the illumination source. The reflection of the display pattern over the object surface is observed with a camera. Each pixel on the display can be seen as an independent single-point light source that can be used to acquire information about a surface point. For example, a fairly "low resolution"  $1000 \times 1000~pix$ display paired with a "low resolution" $1 Mpix$ camera can, in theory, already acquire $1 Mio$ independent 3D surface points. Although these very high numbers are not achievable in practice due to geometrical constraints, our prototype setup experimentally demonstrates measurements with more than 40,000 acquired surface points. Compared to glint tracking with 12 point light sources, this is an improvement in acquired surface points of \textcolor{black}{>$3000 \times$. We demonstrate the feasibility of our approach with quantitative real world experiments: Our measurements on a realistic model eye achieve experimentally evaluated gaze errors between only $0.03^\circ$ and $ 0.12^\circ$ (mean value: $0.09^\circ$)  and precision values between $0.02^\circ$ and $ 0.08^\circ$ (mean value: $0.05^\circ$), which surpasses the current state of the art in glint tracking. Moreover, we show the first quantitative evaluation of our method on real human eyes in vivo of three human subjects. Although early stage, our results already match or approach the best accuracy values of competing methods \cite{adhanom2023eye}. Our measurements display accuracy values between $0.46^\circ$ and $ 0.97^\circ$ (mean value: $0.71^\circ$)  and  precision values between $0.07^\circ$ and $ 0.59^\circ$ (mean value: $0.30^\circ$). Finally, we experimentally verify that the accuracy and precision of our method indeed improve with the number of captured surface points, which leads to interesting conclusions for future advancements in deflectometry-based eye tracking.}

Besides accuracy and precision, another important factor in eye tracking is \textit{speed}. Standard PMD principles normally rely on temporal phase shifting, which requires a temporal sequence of multiple camera images to calculate a single 3D view \cite{huang2018review}. For fast object movements (like those of human eyes), such a temporal phase-shifting procedure is commonly not motion-robust and \textcolor{black}{even a fast execution} would lead to motion artifacts in the final result. For this reason, we exploit a method to measure the complete 3D information about the human eye \textit{in single-shot}. Our implemented approach is based on 2D continuous wavelet transforms for phase measurement \cite{liang2020using}. Due to its single-shot character, it allows for fully motion-robust measurements, regardless of the moving speed of the eye.
We summarize the contributions of our paper as follows:

\begin{itemize}
  
  \item We introduce a new reflection-based method for eye tracking that is fast, accurate, and precise.  Our method exploits dense 3D surface information of the eye (such as surface shape and normals) to estimate the gaze. It does not rely on secondary eye features (such as pupil diameter or veins) and does not require a rendered "eye base model".

\item We develop a novel eye-tracking prototype setup based on single-shot stereo deflectometry~\cite{liang2020using} to capture dense and precise surface information within each camera frame. To the best of our knowledge, this is the first time that deflectometry has been used for eye tracking.

  \item We introduce a novel calibration approach for our prototype system. The approach stands out by its flexibility and ease of use. Its potential to improve other deflectometry systems goes far beyond the application scope of this paper.

  \item We demonstrate the feasibility of our approach with quantitative evaluations \textcolor{black}{on a realistic eye model ``under ideal conditions''. Our evaluated gaze errors are between only $0.03^\circ$ and $ 0.12^\circ$ (mean value: $0.09^\circ$) with precision values between $0.02^\circ$ and $ 0.08^\circ$ (mean value: $0.05^\circ$).} 

  \item \textcolor{black}{We additionally perform experiments on real human eyes in vivo. Although third-party error sources like head movement or target calibration might have been introduced, we achieve gaze accuracy values between $0.46^\circ$ and $ 0.97^\circ$ (mean value: $0.71^\circ$)  and precision values between $0.07^\circ$ and $ 0.59^\circ$ (mean value: $0.30^\circ$).}

   \item \textcolor{black}{We experimentally verify that the accuracy and precision of our method indeed improve with the number of acquired surface points. This result lines out interesting conclusions about future developments that will be discussed in sec.~\ref{3.3}}

\end{itemize}

\begin{figure}[t!]
\centering\includegraphics[width=\linewidth]
{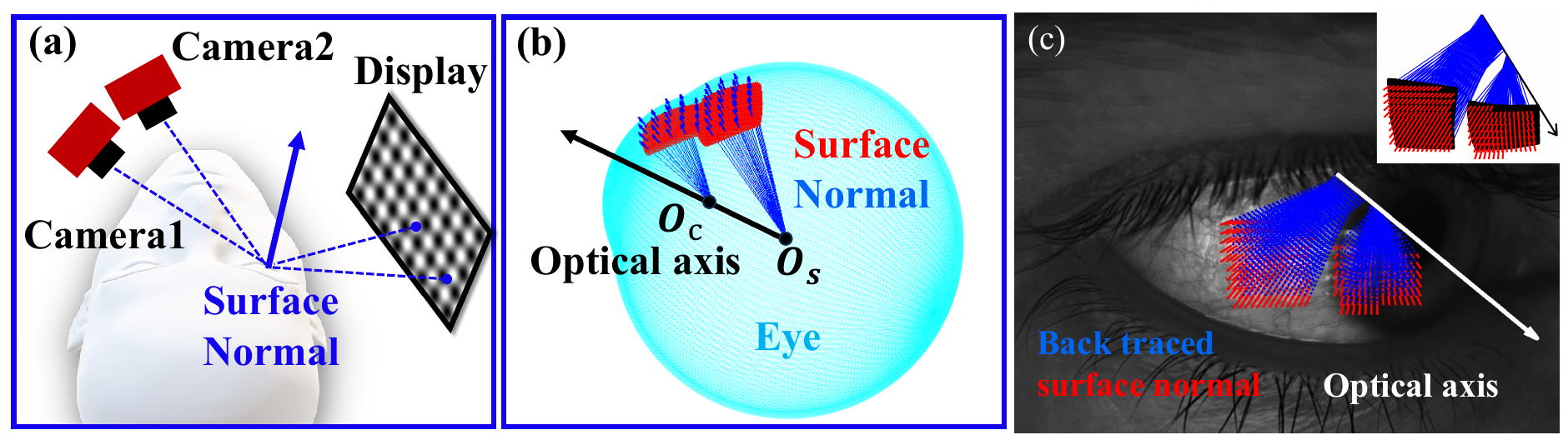}
\caption{\textbf{Deflectometry-based eye-tracking} (a) Schematic of our single-shot deflectometry-based gaze estimation approach: Two cameras observe the specular reflection of a display over the eye surface. Eye surface shape and normal information are calculated from the deformation of the pattern in the camera images.  (b) \textcolor{black}{Simulation: For a perfectly spherical cornea and sclera and no noise, cornea and sclera center can be obtained by tracing the measured surface normals back towards the eye center. The cornea and sclera center points $O_c, O_s$ can be used to obtain the} optical axis and gaze direction.  (c) \textcolor{black}{Real experiment: If cornea and sclera are not spherical, but rotational symmetric, all back traced normals still intersect along the optical axis.}}
\label{title}
\end{figure}

\section{Methods}

Our proposed eye-tracking method leverages accurate and dense 3D measurements of the eye surface. We employ a specifically tailored deflectometry setup to measure the shape and normal map of the eye surface: Instead of using sparse point lights ("glints"), we illuminate the eye with an extended display. Each pixel on the display serves as an individual point light source, and the observed reflection of the display over the eye surface covers both the cornea and sclera region of the eye (see Fig.~\ref{title}b and c). To solve the inherent normal-depth ambiguity\cite{knauer2004phase} and to further enlarge the \textcolor{black}{effective} measured surface area (see Fig.~\ref{title}), we implement a second camera ("stereo-deflectometry").
To facilitate the potential for low latency measurements, we use a single-shot deflectometry approach that exploits one single cross-sinusoid pattern (Fig.~\ref{title}a and Fig.~\ref{fig:pmd}b) rather than classical phase-shifting deflectometry that requires multiple sequentially captured patterns (Fig.~\ref{fig:pmd}e-h). \textcolor{black}{ From the single-shot stereo deflectometry measurements, we retrieve} the depth map and normal map of the eye surface \textcolor{black}{with a novel deflectometry normal and shape reconstruction technique that we have specifically developed to perform under the given geometrical conditions of the human eye (see sec.~2.2). The shape reconstruction goes hand in hand with our gaze angle reconstruction, that leverages geometrical properties about the eye surface to estimate the viewing direction. The basic idea is as follows:}  Given the \textcolor{black}{geometrical} assumption that the shape of the cornea and sclera can be represented by two spherical surfaces with different radii, \textcolor{black}{all} measured surface normals \textcolor{black}{which are back-traced} towards the center of the eye \textcolor{black}{will} aggregate at two points: The center of the cornea sphere and the center of the sclera sphere. The optical axis of the eye is the vector that connects these two points (see Fig.~\ref{title}b). \textcolor{black}{For ``real eyes'' which might not satisfy the 2-sphere assumption but still show rotational symmetry, all backtraced normals still intersect along the optical axis of the eye but do not aggregate at the two distinct center points (see Fig.~\ref{title}c).} \\

In the following, we will provide a detailed explanation of our applied procedures outlined above. We will start with classical phase-shifting deflectometry \textcolor{black}{(sec.~\ref{pmd})} and single-shot deflectometry \textcolor{black}{(sec.~\ref{cwt})}. \textcolor{black}{After briefly explaining the common stereo-deflectometry depth and normal reconstruction procedure in sec.~\ref{sec:recon}, we will introduce our novel reconstruction method in sec.~\ref{sec:gaze}. Our method is able to reconstruct the gaze angle from the initial surface shape and normal deflectometry measurements, exploiting only basic geometrical assumptions about the eye shape. Eventually, we will introduce a novel system calibration method that significantly improves the quality of our measurements and simplifies the calibration process (sec.~\ref{sec:calib})}.

\subsection{Deflectometry} \label{sec:PMD}

\begin{figure}[b!]
\centering\includegraphics[width=13cm]{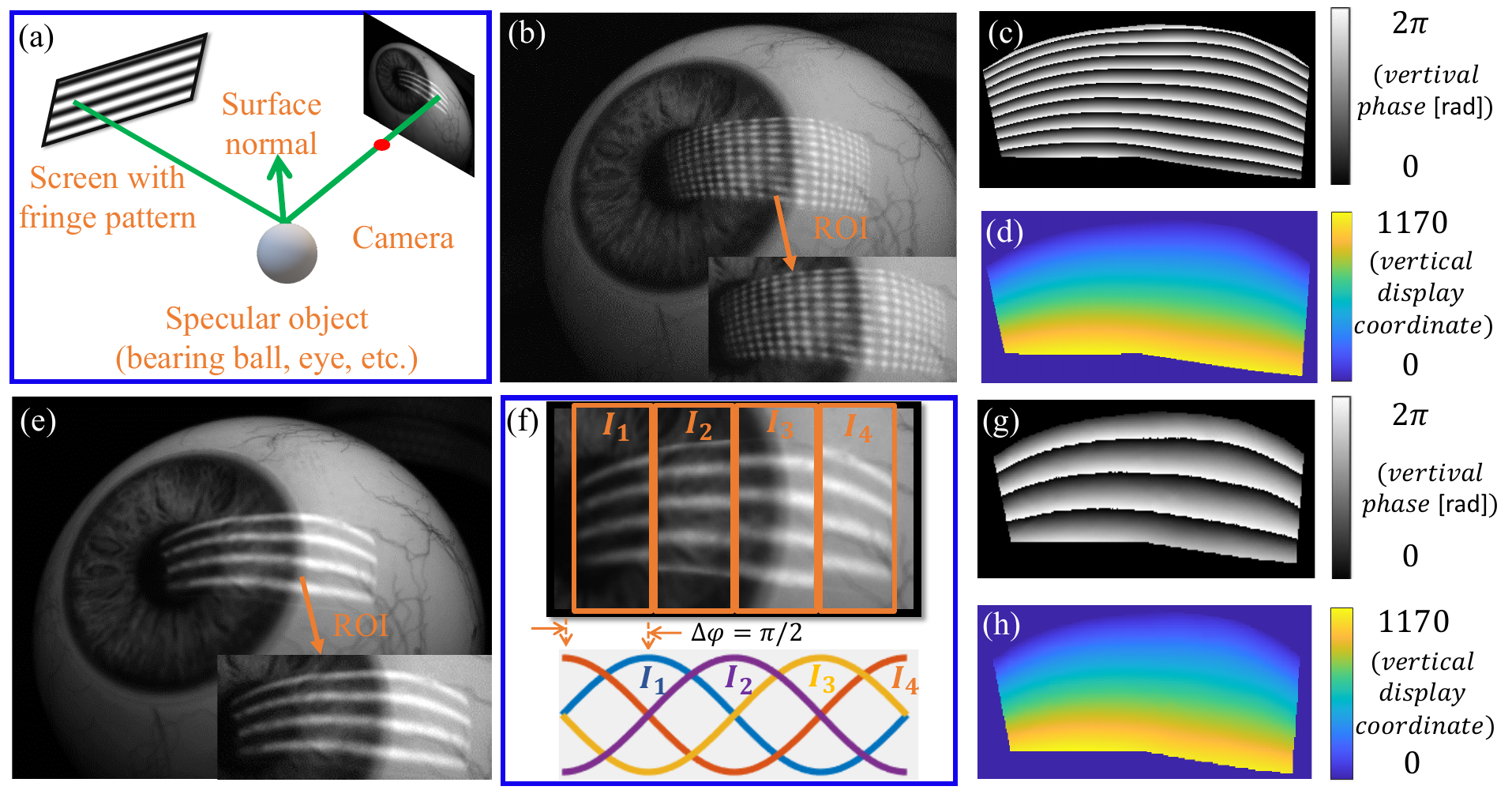}
\caption{ \textbf{Single-shot and multi-shot deflectometry for eye surface measurement.} (a)~Schematic of standard deflectometry: a display with a known pattern illuminates the specular object. The camera observes the deformation of the pattern after reflection. (b) Single-shot deflectometry raw data: Realistic eye model with the reflection of the cross-sinusoid pattern.
(c) and (d) Respective retrieved vertical phase map and unwrapped display-camera correspondence map. (e) Multi-shot deflectometry raw data for comparison: Realistic eye model with reflection of a sinusoidal pattern with horizontal stripes. (f) The four-phase shifting method can be used to retrieve the phase map (g) and eventually the display-camera correspondence map (h) after unwrapping.}
\label{fig:pmd}
\end{figure}

\noindent Deflectometry is an established reflection-based method to reconstruct the surface of specular objects. In a classical deflectometry setup (Fig.~\ref{fig:pmd}a), a camera observes a display over the specular object surface. Depending on the shape of the object surface, the image of the known display pattern is deformed in the camera image. From this deformation, 3D information about the object can be calculated.  The deformation of the pattern is commonly quantified by identifying so-called corresponding points between the display image and the camera image. Eventually, rays are traced between a display pixel and its corresponding camera pixel to calculate the surface normals (and later the shape via integration). As each display pixel emits light in all directions, one single pair of corresponding points leads to many possible combinations of surface normal and surface height. A common solution to this ``depth-normal ambiguity problem''\cite{huang2018review} is to add a second camera and turn the system into a ``stereo deflectometry'' system\cite{knauer2004phase}.

A crucial step in the deflectometry process is to obtain the correspondence between the display and camera pixels. The common procedure is to encode the display pixel positions in a pattern, and eventually decode them on the camera chip by analyzing the pattern. This is commonly done over the phase of displayed sinusoidal intensity patterns of known frequency. \\

\noindent
\subsubsection{Correspondence evaluation in classical phase measuring deflectometry (PMD)}
\label{pmd}
In PMD, the phase values for correspondence evaluation are obtained by temporally shifting vertical and horizontal sinusoidal patterns on the display. Patterns with horizontal stripes (as shown in Fig.~\ref{fig:pmd}e) encode the vertical display coordinate \(y_D\), while patterns with vertical stripes encode the horizontal display coordinate \(x_D\). The reflection of the pattern over the object surface is imaged with the camera. The intensity in each camera pixel $(x_c, y_c)$ can be expressed as 

\begin{equation}
I(x_c,y_c) = A(x_c,y_c) + B(x_c,y_c)\cdot cos(\phi_y(x_c,y_c) + \Delta \phi)~~.
\label{eq1}
\end{equation}

$A(x_c,y_c)$ and  $B(x_c,y_c)$ are terms that contain information about the unknown reflectivity of the respective surface point, the bias illumination, and the unknown background illumination at this point. $\phi_y(x_c,y_c)$ is the vertical phase value that is needed to establish correspondence. In classical PMD, $\phi_y(x_c,y_c)$ is retrieved via temporal phase shifting, e.g., by sequentially changing $\Delta \phi$ to $0, \frac{\pi}{2}, \pi, \frac{3 \pi}{2}$ and acquiring an image $I_k(k=1,2,3,4)$ for each of those phase shifts (``four-phaseshift-method") (see Fig.~\ref{fig:pmd}e and f) \cite{knauer2004phase,huang2018review}. Eventually, $\phi_y(x_c,y_c)$ is calculated by 
\begin{equation}
\phi_y(x_c,y_c) = arctan(\frac{I_{4}-I_{2}}{I_{3}-I_{1}}) ~~.
\label{eq2}
\end{equation}

Fig.~\ref{fig:pmd}g shows the retrieved vertical phase map $\phi_y$ in the camera image, which uniquely encodes the corresponding vertical display coordinates (Fig.~\ref{fig:pmd}h) after unwrapping. The horizontal phase map $\phi_x$ and the horizontal display coordinates are retrieved in an analog fashion by phase-shifting vertical sinusoidal patterns. Eventually, the surface normal is calculated from the obtained correspondence between display pixels and camera pixels via ray tracing (see Fig.~\ref{fig:pmd}.a), where height-normal ambiguities are resolved by exploiting information from the second camera, as described in sec.~\ref{sec:recon}.

Multi-phase shift procedures like the discussed four-phaseshift method provide high-quality results but have one severe drawback for the purpose of this paper: \textit{Several sequentially captured camera images} (8 in this example) are required to capture one single 3D model. During this sequence, the object is not allowed to move. As the human eye is in constant fast motion, this method is not feasible for our purpose in practice. For this reason, we utilize a procedure to facilitate high-quality deflectometry measurements in single-shot. \\

\noindent
\subsubsection{Correspondence evaluation via single-shot deflectometry}
\label{cwt}
To obtain the display-camera correspondence information in single-shot, the unidirectional sinusoidal pattern on the display is replaced with a crossed sinusoidal pattern, i.e., one sinusoid in the horizontal direction, overlaid with a sinusoid in the vertical direction (see Fig.~\ref{fig:pmd}b or pattern schematic in Fig.~\ref{title}a). In analogy to Eq. \ref{eq1}, the intensity in each camera pixel $(x_c, y_c)$ can be expressed by the term

\begin{equation}
I(x_c,y_c) = A(x_c,y_c) + B(x_c,y_c)[cos(\phi_x(x_c,y_c))+cos(\phi_y(x_c,y_c))] ~~~,
\end{equation}

which now contains both the horizontal and vertical phase values $\phi_y(x_c,y_c)$, $\phi_x(x_c,y_c)$, needed to establish correspondence. A common method to retrieve these phase values in single-shot exploits single-sideband demodulation in Fourier space\cite{takeda1983fourier,su2001fourier,liang2016single,ballester2022single}. However, as the frequency evaluation with a Fourier transform is a global operation, the respective method suffers limitations of the allowed maximal object bandwidth. A violation of these limitations leads to artifacts in the reconstruction. In our case, these artifacts are caused by the high surface frequencies at the limbus region of the eye. For this reason, we apply a recent method to \textit{locally} retrieve the phase value in single-shot \cite{liang2020using}. The approach is based on 2D continuous wavelet transforms, and is able to locally evaluate the phase of the crossed fringe pattern. This works by convolving the camera image with a morlet wavelet $\Psi_{s,u_x,u_y}$ that can be varied in scale and translation: 

\begin{equation}\label{eq3}
	\Psi_{s,u_x,u_y} (x_c,y_c)= \frac{1}{s\sqrt{\pi f_b}}  \exp(2\pi f_0(\frac{x_c-u_x}{s})i-\frac{1}{f_b}((\frac{x_c-u_x}{s})^2+(\frac{y_c-u_y}{s})^2))
\end{equation}

Here \(s\) is the scale parameter, \(u_x,u_y\) are the translation parameters in both directions, \(f_0\) is the center frequency of the wavelet and \(f_b\) is the bandwidth. As the object shape does not only influence the local frequency of the pattern in the camera image but also its direction, the wavelet additionally needs to be rotated  to find the maximal wavelet modules:
\begin{equation}\label{eq4}
    \theta_m,s_m = \mathop{\arg\max}_{\theta,s\in \mathbf{R}}\lvert\iint_\mathbf{R^2}I(x_c,y_c)\Psi_{\theta,s,u_x,u_y}(x_c,y_c)dx_cdy_c\lvert
\end{equation}

Eventually, the horizontal phase value $\phi_x(x_c,y_c)$ is determined by the phase angle of the complex valued wavelet transform. For the vertical phase value $\phi_y(x_c,y_c)$, the wavelet is rotated by $90^\circ$ and the process is repeated. The retrieved vertical phase map $\phi_y(x_c,y_c)$ is shown in Fig.~\ref{fig:pmd}c and the respective correspondence map is shown in Fig.~\ref{fig:pmd}d.

\subsubsection{\textcolor{black}{Common} surface normal and shape reconstruction \textcolor{black}{approach from stereo-deflectometry}} \label{sec:recon}

After the correspondence between the camera image and the display has been obtained, the eye surface can be reconstructed. As shown in Fig.~\ref{fig:pmd}a, the surface normal can be obtained by tracing rays between two corresponding points over the specular object surface. However, this mapping is not unique. If the object surface changes its distance to the camera along the ray of sight, the same pair of corresponding points would lead to a different surface normal. This problem is commonly referred to as the `depth-normal-ambiguity problem' \cite{huang2018review}. \textcolor{black}{A common solution is to exploit} information from a second camera that is added to the system (``stereo deflectometry'') \cite{knauer2004phase}. The basic idea: Each camera has its own set of possible surface normal-height combinations, but both sets only match for \textit{one} combination, which is the true height and normal of the surface. Calculating every single surface point via stereo deflectometry is computationally expensive, \textcolor{black}{which is the reason why common stereo deflectometry approaches only} reconstruct surface normal and height at a few sparse "anchor points", \textcolor{black}{used to obtain an initial estimate about the surface height, and the final shape is calculated with an iterative slope integration procedure (see, e.g., \cite{slogsnat2009non, huang2018review ,huang2017zonal}). The} 
 first iteration of the surface normal map \textcolor{black}{is calculated in each camera pixel} ``with respect to'' the initial surface height map. This means that the normal in each pixel is calculated by using the respective height value of the initial surface height map. Eventually, the next iteration of the surface height map is calculated \textcolor{black}{in each camera pixel} by integrating the calculated normal map. A second iteration of the normal map is then calculated with respect to the updated height map, and so on. This process is repeated until the calculated iterations of the surface height map and normal map converge, which typically happens after 2-3 iterations. The result is a highly accurate representation of the measured \textcolor{black}{object shape and normal map}. 

 \textcolor{black}{Applying the "standard" reconstruction procedure described above to measurements of the human eye for gaze estimation results in several challenges (explained in the next section). For this reason, we introduce a novel modification of the established reconstruction method.}

\subsection{\textcolor{black}{Novel normal and shape reconstruction method for gaze estimation from deflectometry measurements}} 
\label{sec:gaze}

\textcolor{black}{ As discussed, our deflectometry-based eye-tracking approach leverages the reconstructed shape and normal map information of the measured eye surface to estimate the gaze angle. Our novel reconstruction method is inspired by the common stereo deflectometry  reconstruction described above, but contains important modifications that lead to improved performance under the given conditions on human eyes.} 

\textcolor{black}{ Our reconstruction algorithm can be roughly divided into two steps: An \textit{initial surface reconstruction step} that exploits the common geometric assumption \cite{beymer2003eye,liu2022method,wang2021vr,wang2023accurate} that the human eye can be approximated by two spheres (cornea and sclera) with different radii, and a \textit{refinement step}, that refines the measured surface and gaze based on the actual deflectometry measurements. The goal of the initial surface reconstruction step is to obtain the centers $O_c, O_s$ of the initially assumed cornea sphere and sclera sphere. For an eye that perfectly satisfies the 2-sphere condition, connecting $O_c$ and  $O_s$ would already deliver the optical axis of the eye which is then used to calculate the gaze. In this idealized case and under the assumption of a noise-free measurement, the center points could be directly retrieved from the standard stereo-deflectometry surface and normal reconstruction,  by simply backtracing all captured surface normals towards the eye center. This case is shown on a simulated 2-sphere eye model in Fig.~\ref{title}b. In real-world measurements, however, small deviations from the 2-sphere model and noise in the deflectometry measurements prevent this basic reconstruction method from working effectively. The shape deviations can be caused, e.g., by small veins on top of the sclera surface, or if the overall shape of the cornea and sclera is not perfectly spherical. Noise in the deflectometry measurement can be caused by sensor or photon noise, slight motion artifacts that can still happen despite the single-shot acquisition, or parts of the eye that occlude that pattern, such as eye lashes. In addition, the standard stereo-deflectometry reconstruction severely limits the effective measurement coverage of the eye surface, as only these surface points can be reconstructed that exhibit a reflection signal for both cameras (see also Fig.~\ref{eyemodle}b). Depending on the geometrical constraints of the setup, this can limit the region of measurable surface points to a fairly small patch in the middle. }

\begin{figure}[t!]
\centering\includegraphics[width=0.8\textwidth]{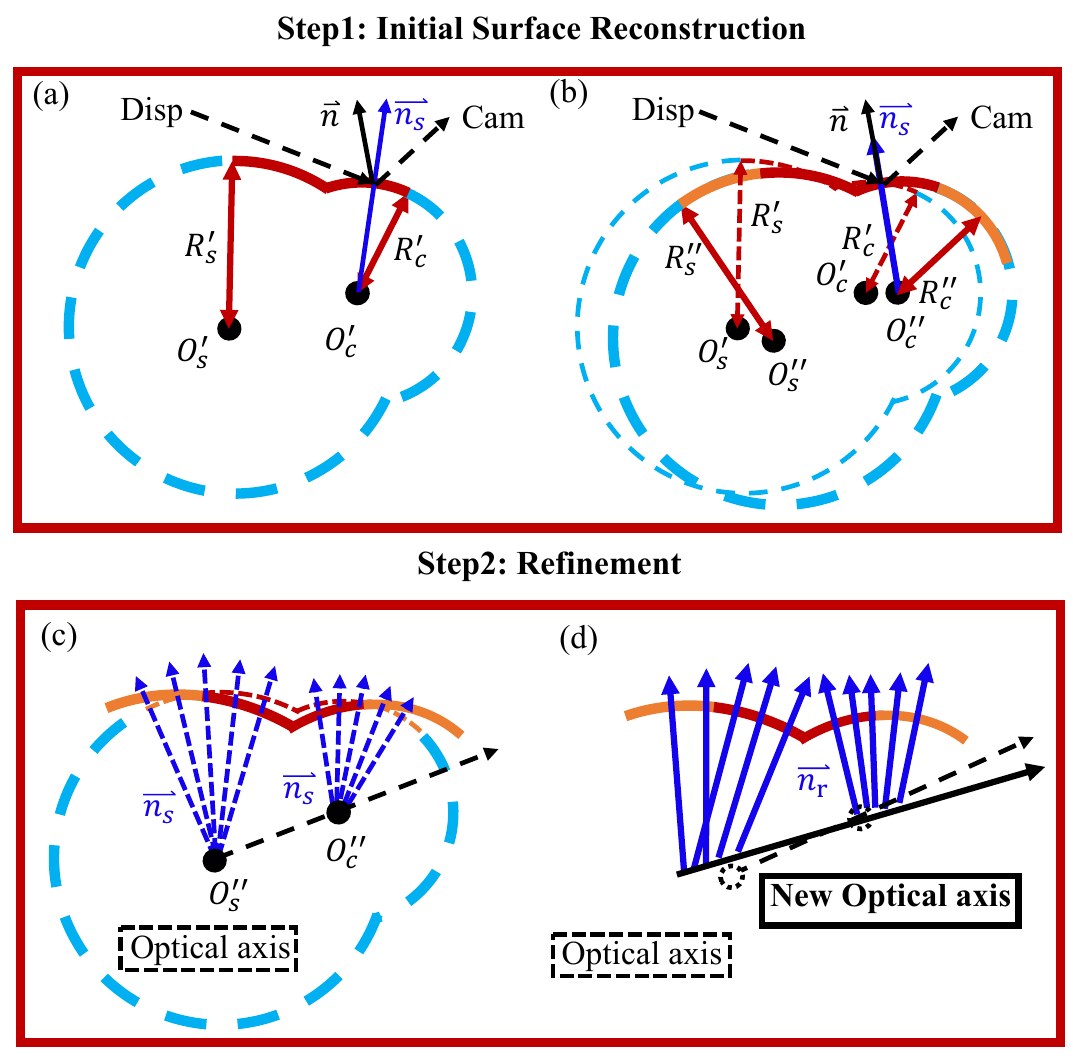}
\caption{\textcolor{black}{\textbf{Novel algorithm for surface reconstruction and gaze estimation.} (a,b) \textit{Initial surface reconstruction:} (a) We use a few points in the initial stereo-deflectometry overlap region to estimate the initial radii $R_c', R_s'$ and centers $O_c', O_s'$ of cornea and sclera to construct an initial 2-sphere model.
 The normals $\overrightarrow{n}$ calculated via deflectometry do not match with the normals $\overrightarrow{n_s}$ of the initial sphere models. (b) We optimize the centers and radii of the initial two-sphere model by expanding the evaluation region to the whole measurement area of both cameras (the red part is the overlap region, orange part is covered by one camera only). New estimates $R_c'', R_s'', O_c'', O_s''$ for radii and centers are obtained by
 minimizing the angular difference between all pairs of $\overrightarrow{n}$ and $\overrightarrow{n_s}$ (see Eq. \eqref{eq:important}). (c,d) \textit{Refinement:} (c) We use the obtained 2-sphere model as initial guess and calculate a new surface representation via iterative deflectometry normal integration\cite{slogsnat2009non, huang2018review ,huang2017zonal}. After this step, the resulting surface parts are not necessarily spherical anymore (in accordance with real human eyes). (d) When back traced, the newly obtained normals do not meet at two distinct center points anymore. However, assuming the human eye is rotationally symmetric, the normals still meet along its optical axis, which is updated accordingly and used to calculate the gaze. } }
\label{pipeline}
\end{figure}

\textcolor{black}{Our novel reconstruction algorithm overcomes these limitations by optimizing the shape and normal reconstruction under the given eye geometry. In step 1 (initial surface reconstruction; see Fig. \ref{pipeline}a and b) we first separate our initial measurement region into cornea and sclera by detecting the abrupt change of fringe deformation at the limbus region. Eventually, we reconstruct surface depth and normal at a few ``anchor points'' (approximately 500 in the shown experiments) in the small overlap region of both cameras (see also Fig.~\ref{eyemodle}b) via the standard stereo-deflectometry reconstruction described in sec.~\ref{sec:recon}. These depth and normal values are then used to obtain an initial guess  $R'_c, R'_s$ of the radii of cornea and sclera as well as their center points $O'_c, O'_s$: The normals are back-traced towards the eye center, and the two aggregation points $O'_c$ and $ O'_s$ are evaluated by detecting best-fit intersection points of all backtraced normals.
We eventually extend the surface reconstruction beyond the boundaries of the overlap region, i.e., to regions where only one camera can detect a reflection signal and where a standard stereo-deflectometry reconstruction would not be possible anymore. This is crucial, as it significantly expands the area where surface points can be measured. We first use our initial guesses $R'_c, R'_s, O'_c, O'_s$ to computationally construct a virtual 2-sphere eye model (see Fig.~\ref{pipeline}a). For each pixel in each camera's effective measurement area $\mathcal{P}$ (including points inside and outside of the overlap region),  we extend the respective camera ray and find the intersection with the surface of the constructed virtual 2-sphere eye model. The surface normal $\overrightarrow{n}$ in each camera pixel is then calculated from the obtained surface intersection point and the corresponding camera display pixel pair (see Fig.~\ref{pipeline}a). An alternative surface normal $\overrightarrow{n_s}$ can be obtained by simply connecting the obtained surface intersection point with the respective center point $O'_c$ or $ O'_s$. As the current rough estimation of the 2-sphere model is not exact,  $\overrightarrow{n}$ and $\overrightarrow{n_s}$ do not overlap (see Fig.~\ref{pipeline}a). We eventually obtain new estimates $O''_c, O''_s, R''_c, R''_s$ by minimizing the angular difference between $\overrightarrow{n}$ and $\overrightarrow{n_s}$ with the following loss objective via gradient decent:}

\begin{equation}
  \mathcal{L} = \frac{1}{|\mathcal{P}|} \sum_{ |\mathcal{P}|} ~\lvert\arccos{(\frac{{\overrightarrow{n}}\cdot{\overrightarrow{n_s}}}{\lvert {\overrightarrow{n}}\rvert \lvert {\overrightarrow{n_s}}\rvert})}\rvert
  \label{eq:important}
\end{equation}

\textcolor{black}{The so-obtained parameters $O''_c, O''_s, R''_c, R''_s$ describe a 2-sphere eye model (shown in Fig.~\ref{pipeline}b) that fits the measured data as close as possible within the limits of the 2-sphere assumption. To accommodate for potential deviations from this assumption, we add an additional "refinement step" in step 2 (see Fig.~\ref{pipeline}c and d): We use the obtained 2-sphere eye model as the initial surface for an iterative integration of the measured normals (see sec.~\ref{sec:recon}). This step slightly changes shape and normals from a 2-sphere model to a representation that is even closer to the real shape (e.g., slightly "egg-shaped"). However, in turn this also means that back traced normals $\overrightarrow{n_r}$ do not exactly intersect at two distinct points $O_c, O_s$ anymore, but rather along the optical axis of the eye, assuming that the eyeball is still rotational symmetric (see also results in Fig.~\ref{eyemodle}c and d). To accommodate for this, we use the line between $O''_c$ and $ O''_s$ as the initial guess of the optical axis and refine its angle and position by minimizing its distance to all back traced surface normals $\overrightarrow{n_r}$ from the integrated eye model. This delivers us our final estimate of the optical axis, which is used to estimate the gaze.}

We emphasize that the retrieved optical axis of the eye is not the visual axis that defines the gaze direction. As the fovea is the spot on the retina that is responsible for the sharpest vision, the actual visual axis slightly differs from the optical axis by an angle $\kappa$. In real-world eye tracking applications, this angle $\kappa$ can be pre-calibrated, e.g., by observing a moving point on a computer display \cite{liu2022method,shih2004novel,zhu2007novel}. \textcolor{black}{Unsurprisingly, the calibration of angle $\kappa$ will bring up additional error sources to the estimated gaze. In our results section, we first evaluate the accuracy of the estimated optical axis by conducting experiments on a realistic eye model mounted on a high-precision rotation stage. Eventually, we conduct experiments on real human eyes in vivo which include the  $\kappa$ angle calibration. We refer to Sec.~\ref{results} for details.}

\subsection{Easy and flexible deflectometry system calibration method} \label{sec:calib}

\begin{figure}[b!]
\centering\includegraphics[width=0.8\linewidth]{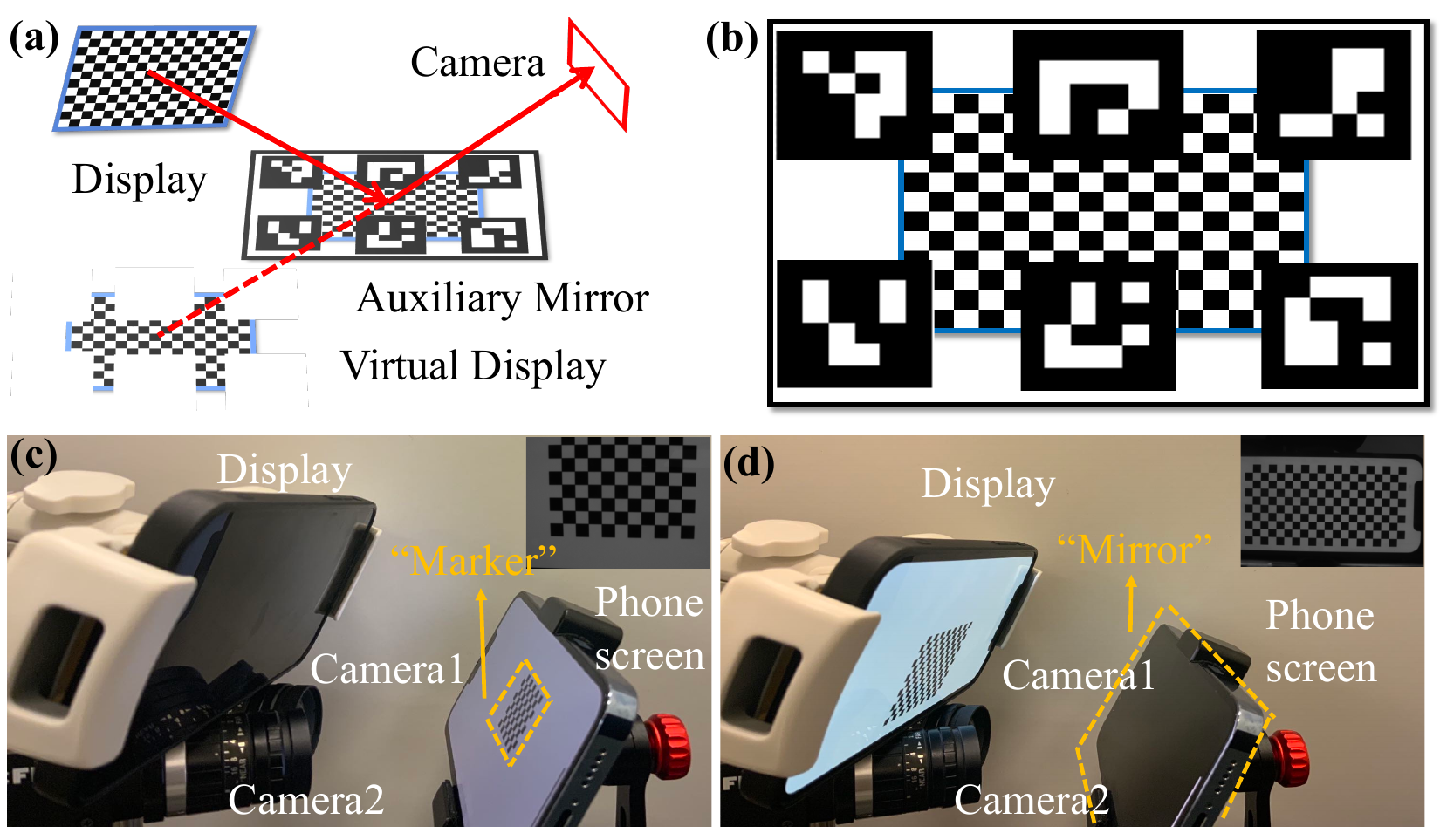}
\caption{\textbf{Novel method for calibration of deflectometry setups} (a) and (b) Shortcomings of the "classical" method\cite{li2021low,wang2022easy}: The display is observed over an auxiliary mirror, whose position in space is determined by attached markers. For precise pose estimation, the markers need to cover a large part of the available FoV, which occludes the display and/or restricts the effective measurement field. (c) and (d) Novel calibration method: A specular phone screen replaces the mirror. In the first step, a dense marker pattern is displayed on the phone screen, which is used to determine the screen's position in space~(c). Eventually, the screen is switched off and functions as a mirror that reflects the display and hence allows for a full-field calibration of the whole display without occlusions (d). \textcolor{black}{The  upper right corner of (c) and (d) are camera images of the  captured ``marker" and ``display reflection''.}}
\label{calib1}
\end{figure}

\noindent The accuracy of deflectometry measurements heavily relies on a high quality system calibration \cite{knauer2004phase,huang2018review,ren2015iterative,wang2022easy}, which includes the intrinsic and extrinsic calibration of all involved cameras, as well as the extrinsic calibration of the entire camera-display system. Zhang's method\cite{zhang2000flexible} is well-established for solving camera intrinsic and extrinsic calibration, where a fixed board with a known pattern (e.g., a checkerboard pattern) is utilized as the calibration target. In practice, the manufacturing quality of the checkerboard highly impacts the calibration result, especially for small field-of-view (FoV) applications. 

During the camera-display extrinsic calibration, the relative pose of both cameras and display is calibrated. As the cameras do not image the display directly, state-of-the-art deflectometry calibration methods use a planar mirror to reflect the display towards the camera (Fig.~\ref{calib1}a).  Each camera sees a virtual image of the display, that can be used to calculate the real display position. The required pose of the mirror can be obtained from markers that are placed on the mirror and observed by each camera. For a highly accurate pose estimation, the markers should be densely distributed across the mirror surface. However, this would occlude parts of the reflected display pattern, leaving specific regions of the display uncalibrated. For this reason, the markers are normally placed at the edge of the mirror, which leads to less accurate calibration results. Moreover, each marker still occupies space in the FoV of each camera which is lost for the actual surface measurement (see Fig.~\ref{calib1}b).

The key idea of our novel calibration method (see Fig~\ref{calib1}c and d) is to replace the static marker-mirror with a switchable specular screen. In the ``on-state'', the screen displays a known calibration pattern (e.g., a checkerboard pattern) that densely fills the entire FoV of both cameras. We evaluate the accurate pose of the screen in space from the respective camera images. Eventually, the screen switches to ``off-state'': No pattern is displayed and the screen acts as a mirror to observe the deflectometry display. This method allows to calibrate the display position without any marker occlusion while simultaneously exploiting the camera's FoV in the most efficient way, which ultimately leads to measurements at much higher accuracy compared to conventional calibration methods\cite{wang2022easy}.

\section{Results}
\label{results}
We demonstrate the feasibility of our approach in real experiments. 
We first quantify the quality of our surface and normal reconstruction method by measuring a spherical bearing ball, which roughly has the same radius as a human eye. Next, we evaluate the \textit{mean relative gaze errors} for fixed rotation positions of a realistic real-world eye model. This experiment is not impacted by additional system errors, \textcolor{black}{ such as head movement of the subjects, or errors in $\kappa$ angle calibration}. The evaluated errors are all $ \leq 0.12^\circ$, with precision values  $ \leq 0.08^\circ$. Moreover, \textcolor{black}{we conduct experiments on a few real human eyes in vivo. Although additional system errors can be involved, our first results still deliver state-of-the-art performance with evaluated gaze accuracies between $0.46^\circ$ and $0.97^\circ$. Moreover, we perform an additional experiment verifying that the accuracy and precision of our method indeed improve with the number of captured surface points.} The following subsections describe our experiments and results in detail.

\subsection{Quantitative evaluation of the surface reconstruction quality}

\begin{figure}[b!]
\centering\includegraphics[width=0.8\linewidth]{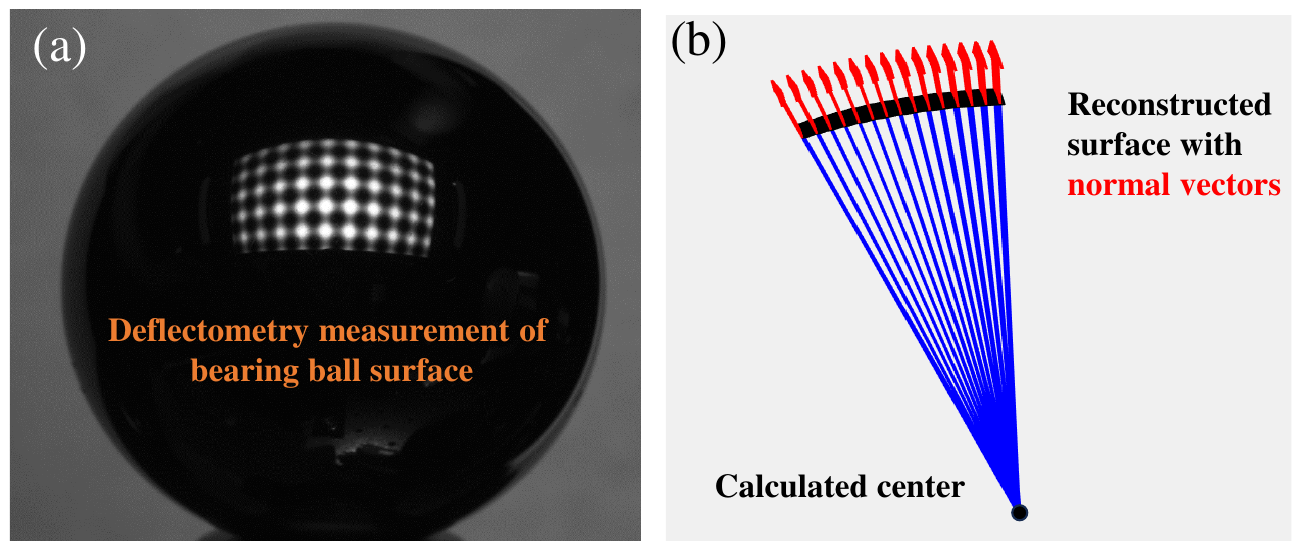}
\caption{\textbf{Quantitative evaluation of surface reconstruction quality.} (a) The system accuracy is evaluated by measuring a spherical bearing ball with the known radius (close to the radius of a real eye). (b) Reconstructed ball surface with measured normal vectors. The calculated radius is 12.02mm, with the ground truth radius being 12mm. The standard deviation of the distance of the calculated sphere center to all normals is  $62\mu m$. }
\label{ball}
\end{figure}

As described above, our method relies on a high-quality deflectometric surface reconstruction for a reliable gaze estimation. For this reason, we first separately evaluate the quality of our surface reconstruction before evaluating the gaze estimation. We use a metallic bearing ball with a well-defined radius that is subject to very little manufacturing tolerances (McMaster-Carr 1598K39, $R= 12mm \pm 0.00125 mm$ ). Our prototype setup to perform all measurements is shown in Fig.~\ref{eyemodle}a. We use two machine vision cameras (Model: flea3 fl3-u3-13s2c) equipped with 9 mm objectives that face the object, which is illuminated by an iPhone 12 pro display (resolution $2532 \times 1170$ pix). The system is calibrated using the novel calibration method described above.

We use the single-shot stereo-deflectometry method \textcolor{black}{paired with our novel surface reconstruction (see sec.~\ref{sec:gaze}, now for only one sphere instead of two)} to measure and reconstruct the spherical surface of the bearing ball. We capture a total of 56,140 surface points and respective surface normals in only one single-shot. An exemplary image from camera 1 is shown in Fig.~\ref{ball}a. The respective surface reconstruction (including normals) is shown in Fig.~\ref{ball}b. To quantify the quality of our reconstruction we calculate two metrics: The \textcolor{black}{best-fit} radius of the \textcolor{black}{final} reconstructed surface \textcolor{black}{after refinement}, and the precision at which the back-traced normals meet at one point (the center of the sphere). The radius of our reconstructed surface is calculated to \textcolor{black}{$R_m =12.02mm$, which deviates only $20\mu m$ from the specified radius of the bearing ball.}  To evaluate the precision of the calculated surface normals, we back-trace all surface normals towards the center of the sphere and estimate the sphere center by calculating their best-fit intersection point. Eventually, the precision $\sigma_m$ is calculated via the standard deviation of the absolute distance of the backtraced normals to the calculated center. We obtain a standard deviation of only \textcolor{black}{$\sigma_m = 62\mu m$}.

\subsection{\textcolor{black}{Experiments on realistic eye model}}
\label{sec:gaze_eval}

Quantifying the performance of our gaze estimation with high reliability requires the design of a systematic experiment. \textcolor{black}{In this regard,} evaluations of gaze directions using real human subjects \textcolor{black}{pose significant design challenges}, mainly for two reasons: (1) one possibility of such an evaluation might require a third-party eye tracker to deliver the ground truth for our measurement. The estimation error of this third-party eye tracker might be higher than the error of our method, which would lead to wrong conclusions. (2) another possible procedure \textcolor{black}{(shown in the next section)}, requires the subject to look at a certain \textcolor{black}{stimulus} on the screen (e.g., a moving point) to calculate the ground truth gaze direction\cite{liu2022method,angelopoulos2021event}. Those methods require very accurate pre-calibration and rely on the assumption that the human subject is not moving during the entire duration of the experiment and also really looks at the signal for most of the time, which introduces another source of error.

\begin{figure}[t!]
\centering\includegraphics[width=0.8\textwidth]{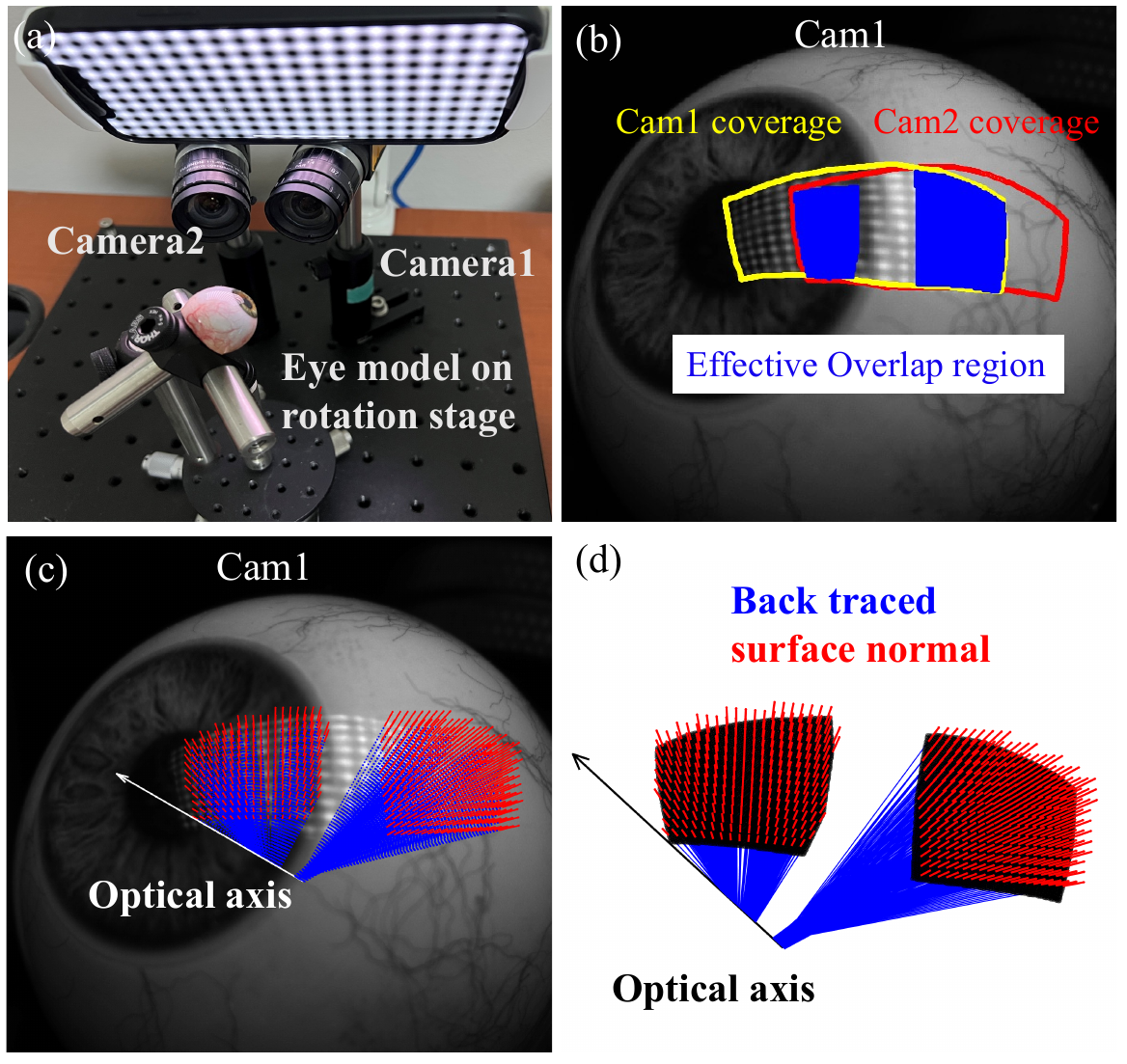}
\caption{\textbf{Quantitative real-world experiments on a realistic eye model.} (a) Prototype setup and eye model mounted on a rotation stage. \textcolor{black}{(b) Overlap and coverage between both cameras, shown in the image taken by camera 1: The yellow frame marks the coverage of camera 1, the red frame marks the coverage of camera 2 back-projected to the image of camera 1. The blue area is the effective overlap region for the initial  stereo deflectometry reconstruction after the ``noisy'' limbus region has been removed  (c) Red arrows are sampled reconstructed surface normals. Blue arrows are back-traced normal vectors that intersect along the optical axis. Results overlaid on captured eye image for visualization. (d) 3D visualization of reconstructed surface and estimated optical axis.}}
\label{eyemodle}
\end{figure}

\textcolor{black}{To first quantify the performance of our eye-tracking method in an isolated way under   ``optimal conditions''}, we designed an experiment that largely eliminates \textcolor{black}{the challenges discussed above}. Instead of a human subject, we use a realistic real-world eye model (with distinct cornea and sclera) that is mounted on a high-precision rotation stage (Model: Thorlabs TTR001, \textcolor{black}{rotation resolution: $0.03^\circ$}). An image of our eye model mounted in our setup is shown in Fig.~\ref{eyemodle}a. We rotate the eye model to several positions of our stage to accurately change the gaze direction by known angles. We emphasize that the ground truth gaze direction for the eye model on the stage is still unknown and can also not be evaluated in a reliable way (see above). For this reason, we define a ``default'' gaze direction by rotating the eye model to $0^\circ$ on the rotation stage and eventually measure the gaze directions as relative angles w.r.t. the default gaze direction. For our evaluation, we rotate the stage to different rotation positions $a$, comprised of $-4^\circ$, $-2^\circ$, $0^\circ$, $2^\circ$, and $4^\circ$. We capture a single-shot measurement at one position and rotate to the next position until we have acquired a total of 20 measurements at each position. On average, we capture  42,673 surface points with respective surface normals in each single-shot measurement at each rotation position. We emphasize that we never capture two consecutive measurements without having rotated the eye model. For each measurement at rotation position $a$ we calculate the relative gazing vector w.r.t. to the ``default'' $0^\circ$ position. Eventually, we calculate the \textit{mean relative error}  $\epsilon_{0^\circ}$ at each rotation angle $a$ as follows: 

\begin{equation}\label{eq:eps}
    \epsilon_{0^\circ}= ||\bar{\theta_{a}} - \bar{\theta_{0^\circ}}| - |a||
\end{equation}

Here $\bar{\theta_{a}}$ is the mean evaluated gaze angle of 20 performed measurements at rotation angle $a$, and $\bar{\theta_{0^\circ}}$ is the mean evaluated gaze angle of 20 performed measurements at rotation angle ${0^\circ}$ (the ``default'' gaze angle). \textcolor{black}{We also emphasize that the evaluation of the \textit{relative} angles avoids the need for a $\kappa$ angle calibration, and $\kappa$ can be simply set to ${0^\circ}$ for this experiment.} The results of \textcolor{black}{our} evaluation are summarized in Tab. \ref{result}. \textcolor{black}{The obtained mean relative gaze errors $\epsilon_{0^\circ}$ are between only $0.03^\circ$ and   $0.12^\circ$ , with the average of all mean relative gaze errors being $0.09^\circ$.}

\begin{table}[h!]
\begin{center}
\begin{tabular}{ | m{3.5cm} | m{0.8cm}| m{0.8cm} |m{0.8cm} |m{0.8cm} |m{0.8cm} |m{0.8cm} | } 
  \hline
  Rotation position $a$ &  -4$^\circ$   & -2$^\circ$  & 0$^\circ$ & 2$^\circ$    & 4$^\circ$ \\ 
  \hline
     Mean relative error $\epsilon_{0^\circ}$ & 0.11$^\circ$ & 0.12$^\circ$ & 0$^\circ$ & 0.03$^\circ$ & 0.10$^\circ$  \\ 
  
  \hline
Precision $\sigma_a$ & 0.08$^\circ$ & 0.06$^\circ$ & 0.04$^\circ$ & 0.07$^\circ$ & 0.02$^\circ$  \\ 
  \hline
\end{tabular}
\end{center}
\caption{Evaluation of the mean relative gaze error and the respective precision values for a real-world measurement of a realistic eye model at different gaze directions.}
\label{result}
\end{table}

In a second step, we evaluate the precision $\sigma_a$ of our method by calculating the standard deviation of our evaluated gazing angles at each rotation position $a$:

\begin{equation}\label{eq:sig}
    \sigma_a = \sqrt{\frac{\sum^{n}_{i=1}(\theta_{a, i} -  \bar{\theta_a})^2}{n}}
\end{equation}

$\theta_{a, i}$ is the  $i^{th}$ measurement at rotation position $a$, and $n = 20$ in this experiment. As also shown in Tab. \ref{result} \textcolor{black}{the gaze precision $\sigma_a$ is evaluated between  $ 0.02^\circ$ and $ 0.08^\circ$, with an average of $ 0.05^\circ$.}

\subsection{\textcolor{black}{Quantitative evaluation on real human eyes in vivo}}
\label{3.3}

\begin{figure}[b!]
    \centering
    \includegraphics[width=\textwidth]{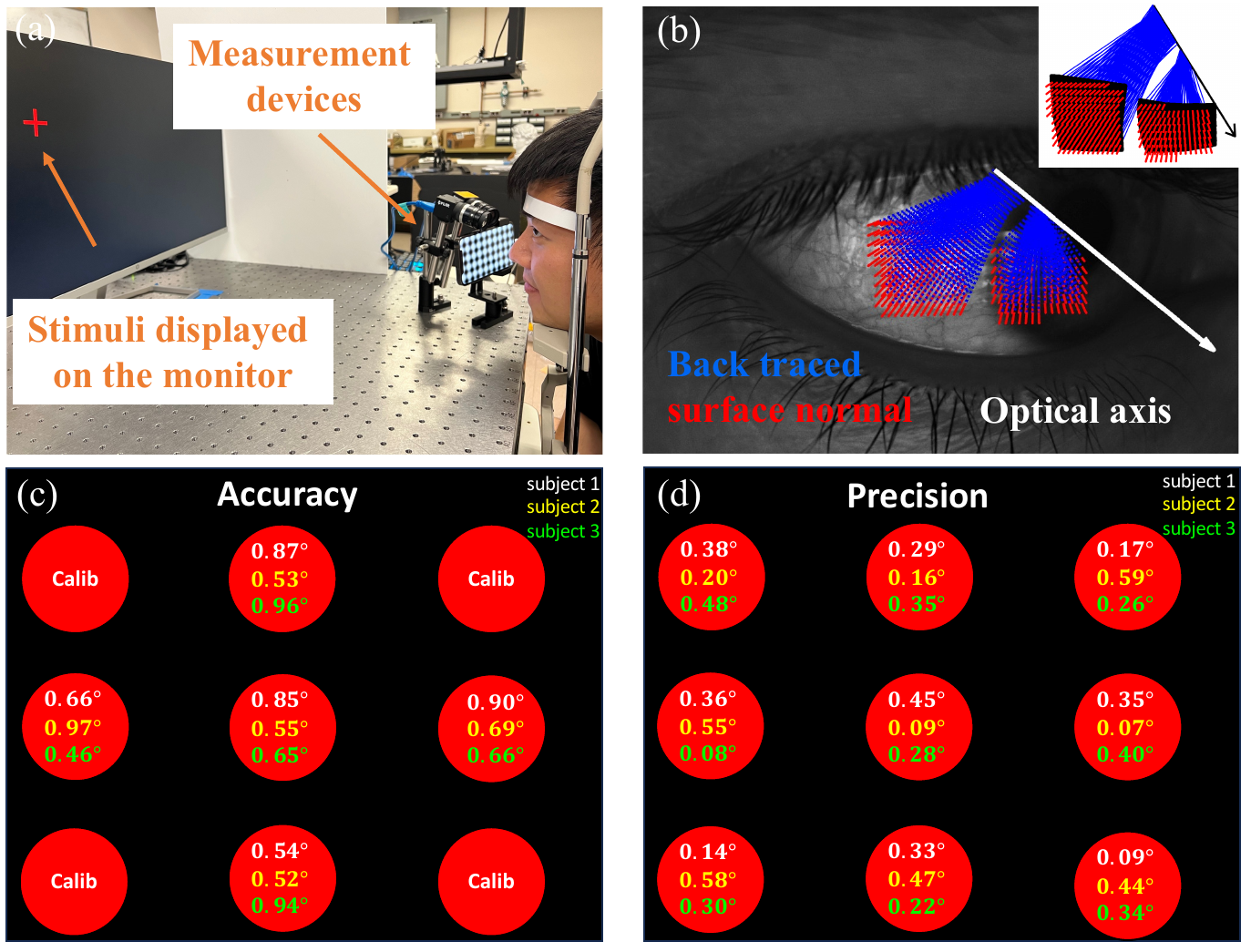}
    \caption{\textcolor{black}{\textbf{Quantitative experiment on real human eyes in vivo} (a) Setup: a total of three human subjects are asked to place their head on a chin rest and stare at stimuli (3x3) on a remote computer screen. 10 measurements are taken for each stimulus position for each subject. (b) Sample image, overlaid with evaluation result. (c)~Evaluated accuracy for each subject at each stimulus position. The four corner stimuli positions are used for $\kappa$-angle and gaze calibration. Achieved accuracy values vary between $0.46^\circ$ and $0.97^\circ$. (d) Evaluated precision for each subject at each stimulus position. Achieved precision values vary between $0.07^\circ$ and $0.59^\circ$.}}
    \label{realeye}
\end{figure}
\textcolor{black}{In addition to the previous experiment which is performed on a realistic eye model under "optimal conditions", we conduct experiments on real human eyes in vivo to demonstrate the feasibility of our method in real eye-tracking tasks. We place our setup on a table and instruct the subjects to rest their heads on a chin rest to stabilize the head pose (see Fig.~\ref{realeye}a). Eventually, we generate stimuli at 9 different positions on a fixed remote monitor and instruct the subjects to fixate on the stimuli. As discussed in, e.g. \cite{ramanauskas2006calibration,harezlak2014towards,angelopoulos2021event,nystrom2013influence}, calibrating the monitor target is crucial for the correct performance of the experiment. We calibrate the remote monitor into our setup using the same method described in Sec.~\ref{sec:calib}, i.e., by observing signals displayed on the monitor with our system cameras via the reflection over a phone screen that we also use as "marker plate". In total, we display 9 stimuli ($3\times3$ grid), which span an angular field of view of $\sim 30^\circ \times 15^\circ$ in azimuth and elevation, respectively. The experiment is performed on three human subjects. For each of these 3 subjects, we capture 10 gaze estimations per stimulus position, using the methodology described above. Eventually, we use the four corner stimuli on the screen (see Fig.~\ref{realeye}c) to perform the optical-visual axes transformation calibration step, necessary to obtain gaze angles in the real-world coordinate system: For each "corner stimulus", we evaluate the mean optical axis (from 10 measurements) in the camera coordinate system and eventually determine the matrix that transforms the results to the respective stimuli positions on the screen via least squares optimization. Subsequently, the obtained matrix (which now already includes the $\kappa$ angle calibration) is applied to all estimated optical axes to evaluate the absolute visual gaze direction in world coordinates. For each stimulus that was not used for calibration, we calculate the RMSE w.r.t. the ground truth gaze angle from all 10 measurements. This delivers us an accuracy value for our method at each of the 5 remaining stimulus positions for each subject (see Fig.~\ref{realeye}c). Moreover, we calculate the precision at all stimulus positions for all subjects by evaluating the RMSE w.r.t. the mean evaluated gaze angle (from 10 measurements, see Fig.~\ref{realeye}d). The obtained accuracy and precision values from all three subjects are shown in Fig.~\ref{realeye}c and d for all stimulus positions.  The evaluated accuracy values for all three subjects vary between $0.46^\circ$ and $0.97^\circ$. The mean accuracy (averaged over all stimuli positions) for each subject is $0.76^\circ$, $0.65^\circ$, and $0.73^\circ$, respectively. The evaluated precision values for all three subjects vary between $0.07^\circ$ and $0.59^\circ$. The mean precision (averaged over all stimuli positions) for each subject is $0.28^\circ$, $0.35^\circ$, and $0.30^\circ$, respectively.}

\textcolor{black}{Our experiments mark the first-ever demonstration and quantitative evaluation of deflectometry-based eye tracking on real human eyes in vivo. Although early stage, our first experiments already reach accuracy and precision values which are on par with the best current state-of-the-art methods. However, the even higher performance of our experiments on the model eye could not be reached for all human eye measurements. We attribute this to the fact that the present experiment required additional calibration steps (which introduce additional errors), potential head movements or gaze fixation errors of the human subjects, and slight surface reflectance variations on the real human eyes that might not be correctly represented by the model eye and could lead to additional noise in the data. We also emphasize that our lowest accuracy and precision values (e.g., subject 3, stimulus middle left: accuracy $0.46^\circ$, precision  $0.08^\circ$) display similar performance to our model eye measurement (although the present experiment evaluates the absolute error instead of the relative error), meaning that the above-mentioned effects are largely neglectable for these ``good'' measurements. Another result worthwhile to mention is that, in contrast to other competing methods,  there is no obvious degradation in performance when comparing measurements at the center stimuli positions to those at the outer positions on the edge of the $\sim 30^\circ \times 15^\circ$  angular field of view - at least not for the shown three initial experiments. Experiments on more human subjects under different conditions will be part of our future work.}

\textcolor{black}{In a last experiment, we test the hypothesis of whether more acquired 3D data points of the eye surface lead to a better performance of our method or not. We exemplarily pick the obtained measurements from subject 1 for our test. The complete data set of one single gaze measurement consists of roughly 40,000 surface point-normal pairs. For each measurement, we artificially thin out the acquired point clouds by different factors, by picking and processing randomly distributed subsets of the acquired data. Eventually, we calculate the mean accuracy and mean precision (i.e., the mean value of the accuracy and precision values from all stimuli positions) for each point cloud density. Our results are shown in Tab. \ref{result3}. It can be seen that both accuracy and precision steadily degrade from $0.76^\circ$ / $0.28^\circ$ for full point cloud density ($\sim 40,000$ points) to $1.32^\circ$ / $1.2^\circ$ for 1/16 point cloud density ($\sim 2,500$ points). The evaluation of both other subjects (not shown here) shows a similar trend. This confirms the hypothesis that a higher number of acquired data points does indeed have a positive effect on the precision and accuracy of our method. The result also leads to interesting conclusions for possible advancements of our method in the future, where we expect that even lower gaze errors can be reached if the number of acquired point-normal pairs and their coverage (i.e., distribution over the eye ball) can be further improved.}

\begin{table}[h!]
\begin{center}
\begin{tabular}{ | m{2.5cm} | m{0.8cm}| m{0.8cm} |m{0.8cm} |m{0.8cm} |m{0.8cm} |m{0.8cm} | } 
  \hline
  Ratio &  full & 1/2   & 1/4  & 1/8 & 1/16    \\ 
  \hline
  Approx points &  40,000 & 20,000   & 10,000  & 5,000 & 2,500    \\ 
  \hline
     Mean accuracy  & 0.76$^\circ$ & 0.82$^\circ$ & 1.27$^\circ$ & 1.35$^\circ$ & 1.32$^\circ$  \\ 
  
  \hline
Mean precision  & 0.28$^\circ$ & 0.42$^\circ$ & 0.67$^\circ$ & 0.89$^\circ$ & 1.20$^\circ$  \\ 
  \hline
\end{tabular}
\end{center}
\caption{Analysis of the influence of data sparseness of subject 1}
\label{result3}
\end{table}

\section{Summary, discussion, and outlook}

\noindent This paper introduced a novel method for reflection-based eye-tracking that exploits dense 3D eye surface reconstructions from single-shot deflectometry measurements. State-of-the-art reflection-based methods (e.g., glint tracking) only sample the eye surface at $\sim 10-12$ points. In our experiments, we can easily acquire >40,000 \textcolor{black}{point-normal pairs} of the eye surface in each single-shot measurement. \textcolor{black}{We demonstrated our novel method via quantitative measurements on a realistic eye model, and on real human eyes in vivo. Our human eye measurements mark the first-ever quantitative deflectometry eye-tracking experiments on human eyes, and our first experiments already reach accuracy and precision values which are on par with the best current state-of-the-art methods.  The achieved accuracy values are all between $0.46^\circ$ and $ 0.97^\circ$, with precision values being between $0.07^\circ$ and $ 0.59^\circ$. Our measurements on the realistic eye model achieve an even better performance, reaching mean relative gaze errors between only $0.03^\circ$ and $ 0.12^\circ$, with precision values between $0.02^\circ$ and $ 0.08^\circ$.}

Besides its single-shot ability and high accuracy, our method has additional interesting attributes that will potentially become important for future \textcolor{black}{eye-tracking} application scenarios: It does not rely on 2D eye texture features such as iris images or pupil center, whose exact position is prone to errors due to refraction of light rays at the cornea surface. 

Another interesting feature comes from the 3D imaging capability of deflectometry: The  ``byproduct'' of our gaze estimation is a very accurate measurement of the eye surface. As described in\cite{liang2016single}, this surface measurement can be used to quantify vision impairments. In a potential VR/AR headset setting, this information can be used to correct the vision impairments of the user ``on the fly", meaning that the user would not be required to wear contact lenses or glasses, and a previous calibration is not required.

\textcolor{black}{Our method is of course also not without limitations. In our initial surface reconstruction step (before the refinement step, see sec.~\ref{sec:gaze}) we assume that  cornea and sclera are spherical. This is a common assumption in eye tracking \cite{beymer2003eye,liu2022method,wang2021vr,wang2023accurate}, but it makes the initial surface reconstruction susceptible to errors if very unusual eye shapes are measured. Although we mitigate this model assumption in the second ``refinement step'' of the reconstruction, it might be worth exploring more sophisticated eye models \cite{corbett1994topography,atchison2004eye} for the initial surface reconstruction step in the future.}

\textcolor{black}{Although we have shown that our gaze estimation results improve with a larger number of acquired surface point-normal pairs, we believe that a larger measurement field could also have a positive effect and could further improve our results. Increasing the effective measurement field (i.e., the coverage/size of the display reflection) represents a challenge, however, as the field restriction is part of an inherent limitation of deflectometry \cite{burke2023deflectometry}. Potential solutions would require larger and/or curved displays, which would become unfeasible for certain applications. In our future work, we seek to exploit display-free solutions, such as fixed known back-lit patterns that could cover the interior of VR glasses, or exploiting the known content of the main (VR) display (e.g., frames of a movie, game, etc.) itself as a ``pattern'' for our deflectometry measurement.}

\textcolor{black}{Another limitation to mention is the higher system calibration effort compared to standard image-based methods which typically only require a calibration of the used camera.}

\textcolor{black}{Similar to other eye tracking methods, our method also has a ``dead range,'' where reliable results can not be obtained anymore. For many glint tracking methods, this dead range is reached for larger gaze angles, when glints are not located on the cornea anymore and have moved to the sclera. In our case, the dead range is reached if the display reflection does not cover both the cornea and sclera anymore or when the reflection on the sclera is largely occluded (e.g., by upper or lower eyelids). Similar to the discussed coverage/measurement field limitation, these effects are strongly dependent on the geometry of our setup (display size and distance, display shape, component placement, etc.), and could potentially be improved in a similar fashion as discussed above.}

\textcolor{black}{ Besides the mentioned potential future improvements,} we will also seek to further improve our method by replacing certain steps in our pipeline with data-driven approaches, and also by combining it with certain aspects
of our inverse rendering-based technique\cite{wang2023optimization,cossairt2020low}. 
We hope that our introduced method will become part of a new wave of accurate and fast eye-tracking methods that enable unprecedented features in VR/AR development and drastically improve prediction accuracy in clinical use and psychology research. \\


\bibliography{reference}

\begin{thebibliography}{10}
\newcommand{\enquote}[1]{``#1''}

\bibitem{holmqvist2011eye}
K.~Holmqvist, R.~Andersson, R.~Dewhurst, H.~Jarodzka, J.~Van~de Weijer \emph{et~al.}, \emph{Eye tracking: A comprehensive guide to methods and measures} (oup Oxford, 2011).

\bibitem{adhanom2023eye}
I.~B. Adhanom, P.~MacNeilage, and E.~Folmer, \enquote{Eye tracking in virtual reality: a broad review of applications and challenges,} {\protect\JournalTitle{Virtual Reality}} pp. 1--24 (2023).

\bibitem{clark2019potential}
R.~Clark, J.~Blundell, M.~J. Dunn, J.~T. Erichsen, M.~E. Giardini, I.~Gottlob, C.~Harris, H.~Lee, L.~Mcilreavy, A.~Olson \emph{et~al.}, \enquote{The potential and value of objective eye tracking in the ophthalmology clinic,} {\protect\JournalTitle{Eye}} \textbf{33}, 1200--1202 (2019).

\bibitem{merali2019eye}
N.~Merali, D.~Veeramootoo, and S.~Singh, \enquote{Eye-tracking technology in surgical training,} {\protect\JournalTitle{Journal of Investigative Surgery}} \textbf{32}, 587--593 (2019).

\bibitem{rahal2019understanding}
R.-M. Rahal and S.~Fiedler, \enquote{Understanding cognitive and affective mechanisms in social psychology through eye-tracking,} {\protect\JournalTitle{Journal of Experimental Social Psychology}} \textbf{85}, 103842 (2019).

\bibitem{hessels2019eye}
R.~S. Hessels and I.~T. Hooge, \enquote{Eye tracking in developmental cognitive neuroscience--the good, the bad and the ugly,} {\protect\JournalTitle{Developmental cognitive neuroscience}} \textbf{40}, 100710 (2019).

\bibitem{nishino2004world}
K.~Nishino and S.~K. Nayar, \enquote{The world in an eye [eye image interpretation],} in \emph{Proceedings of the 2004 IEEE Computer Society Conference on Computer Vision and Pattern Recognition, 2004. CVPR 2004.},  vol.~1 (IEEE, 2004), pp. I--I.

\bibitem{lu2011head}
F.~Lu, T.~Okabe, Y.~Sugano, and Y.~Sato, \enquote{A head pose-free approach for appearance-based gaze estimation.} in \emph{BMVC,}  (2011), pp. 1--11.

\bibitem{lu2011inferring}
F.~Lu, Y.~Sugano, T.~Okabe, and Y.~Sato, \enquote{Inferring human gaze from appearance via adaptive linear regression,} in \emph{2011 International Conference on Computer Vision,}  (IEEE, 2011), pp. 153--160.

\bibitem{li2015gaze}
J.~Li and S.~Li, \enquote{Gaze estimation from color image based on the eye model with known head pose,} {\protect\JournalTitle{IEEE Transactions on Human-Machine Systems}} \textbf{46}, 414--423 (2015).

\bibitem{valliappan2020accelerating}
N.~Valliappan, N.~Dai, E.~Steinberg, J.~He, K.~Rogers, V.~Ramachandran, P.~Xu, M.~Shojaeizadeh, L.~Guo, K.~Kohlhoff \emph{et~al.}, \enquote{Accelerating eye movement research via accurate and affordable smartphone eye tracking,} {\protect\JournalTitle{Nature communications}} \textbf{11}, 1--12 (2020).

\bibitem{coutinho2006free}
F.~L. Coutinho and C.~H. Morimoto, \enquote{Free head motion eye gaze tracking using a single camera and multiple light sources,} in \emph{2006 19th Brazilian Symposium on Computer Graphics and Image Processing,}  (IEEE, 2006), pp. 171--178.

\bibitem{hennessey2009improving}
C.~A. Hennessey and P.~D. Lawrence, \enquote{Improving the accuracy and reliability of remote system-calibration-free eye-gaze tracking,} {\protect\JournalTitle{IEEE transactions on biomedical engineering}} \textbf{56}, 1891--1900 (2009).

\bibitem{mestre2018robust}
C.~Mestre, J.~Gautier, and J.~Pujol, \enquote{Robust eye tracking based on multiple corneal reflections for clinical applications,} {\protect\JournalTitle{Journal of biomedical optics}} \textbf{23}, 035001 (2018).

\bibitem{liu2022method}
J.~Liu, J.~Chi, and S.~Fan, \enquote{A method for accurate 3d gaze estimation with a single camera and two collinear light sources,} {\protect\JournalTitle{IEEE Transactions on Instrumentation and Measurement}}  (2022).

\bibitem{beymer2003eye}
D.~Beymer and M.~Flickner, \enquote{Eye gaze tracking using an active stereo head,} in \emph{2003 IEEE Computer Society Conference on Computer Vision and Pattern Recognition, 2003. Proceedings.},  vol.~2 (IEEE, 2003), pp. II--451.

\bibitem{shih2004novel}
S.-W. Shih and J.~Liu, \enquote{A novel approach to 3-d gaze tracking using stereo cameras,} {\protect\JournalTitle{IEEE Transactions on Systems, Man, and Cybernetics, Part B (Cybernetics)}} \textbf{34}, 234--245 (2004).

\bibitem{tan2002appearance}
K.-H. Tan, D.~J. Kriegman, and N.~Ahuja, \enquote{Appearance-based eye gaze estimation,} in \emph{Sixth IEEE Workshop on Applications of Computer Vision, 2002.(WACV 2002). Proceedings.},  (IEEE, 2002), pp. 191--195.

\bibitem{baluja1993non}
S.~Baluja and D.~Pomerleau, \enquote{Non-intrusive gaze tracking using artificial neural networks,} {\protect\JournalTitle{Advances in Neural Information Processing Systems}} \textbf{6} (1993).

\bibitem{krafka2016eye}
K.~Krafka, A.~Khosla, P.~Kellnhofer, H.~Kannan, S.~Bhandarkar, W.~Matusik, and A.~Torralba, \enquote{Eye tracking for everyone,} in \emph{Proceedings of the IEEE conference on computer vision and pattern recognition,}  (2016), pp. 2176--2184.

\bibitem{zhu2017monocular}
W.~Zhu and H.~Deng, \enquote{Monocular free-head 3d gaze tracking with deep learning and geometry constraints,} in \emph{Proceedings of the IEEE International Conference on Computer Vision,}  (2017), pp. 3143--3152.

\bibitem{patel2019refractive}
S.~Patel and L.~Tutchenko, \enquote{The refractive index of the human cornea: A review,} {\protect\JournalTitle{Contact Lens and Anterior Eye}} \textbf{42}, 575--580 (2019).

\bibitem{zhu2007novel}
Z.~Zhu and Q.~Ji, \enquote{Novel eye gaze tracking techniques under natural head movement,} {\protect\JournalTitle{IEEE Transactions on biomedical engineering}} \textbf{54}, 2246--2260 (2007).

\bibitem{hansen2009eye}
D.~W. Hansen and Q.~Ji, \enquote{In the eye of the beholder: A survey of models for eyes and gaze,} {\protect\JournalTitle{IEEE transactions on pattern analysis and machine intelligence}} \textbf{32}, 478--500 (2009).

\bibitem{kar2017review}
A.~Kar and P.~Corcoran, \enquote{A review and analysis of eye-gaze estimation systems, algorithms and performance evaluation methods in consumer platforms,} {\protect\JournalTitle{IEEE Access}} \textbf{5}, 16495--16519 (2017).

\bibitem{zhao2023retinal}
Y.~Zhao, D.~Lindberg, B.~Cleary, O.~Mercier, R.~Mcclelland, E.~Penner, Y.-J. Lin, J.~Majors, and D.~Lanman, \enquote{Retinal-resolution varifocal vr,} in \emph{ACM SIGGRAPH 2023 Emerging Technologies,}  (2023), pp. 1--3.

\bibitem{knauer2004phase}
M.~C. Knauer, J.~Kaminski, and G.~Häusler, \enquote{Phase measuring deflectometry: a new approach to measure specular free-form surfaces,} in \emph{Optical Metrology in Production Engineering,}  vol. 5457 (SPIE, 2004), pp. 366--376.

\bibitem{knauer2008measuring}
M.~C. Knauer, C.~Richter, P.~Vogt, and G.~H{\"a}usler, \enquote{Measuring the refractive power with deflectometry in transmission,} {\protect\JournalTitle{DGaO Proc}}  (2008).

\bibitem{hofer2016infrared}
S.~H{\"o}fer, J.~Burke, and M.~Heizmann, \enquote{Infrared deflectometry for the inspection of diffusely specular surfaces,} {\protect\JournalTitle{Advanced Optical Technologies}} \textbf{5}, 377--387 (2016).

\bibitem{faber2012deflectometry}
C.~Faber, E.~Olesch, R.~Krobot, and G.~H{\"a}usler, \enquote{Deflectometry challenges interferometry: the competition gets tougher!} in \emph{Interferometry XVI: techniques and analysis,}  vol. 8493 (SPIE, 2012), pp. 232--246.

\bibitem{olesch2014deflectometric}
E.~Olesch, G.~H{\"a}usler, A.~W{\"o}rnlein, F.~Stinzing, and C.~van Eldik, \enquote{Deflectometric measurement of large mirrors,} {\protect\JournalTitle{Advanced Optical Technologies}} \textbf{3}, 335--343 (2014).

\bibitem{hausler2013deflectometry}
G.~H{\"a}usler, C.~Faber, E.~Olesch, and S.~Ettl, \enquote{Deflectometry vs. interferometry,} in \emph{Optical measurement systems for industrial inspection VIII,}  vol. 8788 (SPIE, 2013), pp. 367--377.

\bibitem{burke2023deflectometry}
J.~Burke, A.~Pak, S.~H{\"o}fer, M.~Ziebarth, M.~Roschani, and J.~Beyerer, \enquote{Deflectometry for specular surfaces: an overview,} {\protect\JournalTitle{Advanced Optical Technologies}} \textbf{12}, 1237687 (2023).

\bibitem{su2012scots}
P.~Su, S.~Wang, M.~Khreishi, Y.~Wang, T.~Su, P.~Zhou, R.~E. Parks, K.~Law, M.~Rascon, T.~Zobrist \emph{et~al.}, \enquote{Scots: a reverse hartmann test with high dynamic range for giant magellan telescope primary mirror segments,} in \emph{Modern Technologies in Space-and Ground-based Telescopes and Instrumentation II,}  vol. 8450 (SPIE, 2012), pp. 332--340.

\bibitem{trumper2016instantaneous}
I.~Trumper, H.~Choi, and D.~W. Kim, \enquote{Instantaneous phase shifting deflectometry,} {\protect\JournalTitle{Optics express}} \textbf{24}, 27993--28007 (2016).

\bibitem{willomitzer2020hand}
F.~Willomitzer, C.-K. Yeh, V.~Gupta, W.~Spies, F.~Schiffers, A.~Katsaggelos, M.~Walton, and O.~Cossairt, \enquote{Hand-guided qualitative deflectometry with a mobile device,} {\protect\JournalTitle{Optics express}} \textbf{28}, 9027--9038 (2020).

\bibitem{li2021low}
Y.~Li, C.-K. Yeh, B.~Xu, F.~Schiffers, M.~Walton, J.~Tumblin, A.~Katsaggelos, F.~Willomitzer, and O.~Cossairt, \enquote{A low-cost solution for 3d reconstruction of large-scale specular objects,} in \emph{Computational Optical Sensing and Imaging,}  (Optica Publishing Group, 2021), pp. CW4H--3.

\bibitem{liang2016single}
H.~Liang, E.~Olesch, Z.~Yang, and G.~H{\"a}usler, \enquote{Single-shot phase-measuring deflectometry for cornea measurement,} {\protect\JournalTitle{Advanced Optical Technologies}} \textbf{5}, 433--438 (2016).

\bibitem{zheng2023fringe}
Y.~Zheng, Q.~Chao, Y.~An, S.~Hirsh, and A.~Fix, \enquote{Fringe projection-based single-shot 3d eye tracking using deep learning and computer graphics,} in \emph{Optical Architectures for Displays and Sensing in Augmented, Virtual, and Mixed Reality (AR, VR, MR) IV,}  vol. 12449 (SPIE, 2023), pp. 265--275.

\bibitem{huang2018review}
L.~Huang, M.~Idir, C.~Zuo, and A.~Asundi, \enquote{Review of phase measuring deflectometry,} {\protect\JournalTitle{Optics and Lasers in Engineering}} \textbf{107}, 247--257 (2018).

\bibitem{liang2020using}
H.~Liang, T.~Sauer, and C.~Faber, \enquote{Using wavelet transform to evaluate single-shot phase measuring deflectometry data,} in \emph{Applications of Digital Image Processing XLIII,}  vol. 11510 (SPIE, 2020), pp. 404--410.

\bibitem{takeda1983fourier}
M.~Takeda and K.~Mutoh, \enquote{Fourier transform profilometry for the automatic measurement of 3-d object shapes,} {\protect\JournalTitle{Applied optics}} \textbf{22}, 3977--3982 (1983).

\bibitem{su2001fourier}
X.~Su and W.~Chen, \enquote{Fourier transform profilometry:: a review,} {\protect\JournalTitle{Optics and lasers in Engineering}} \textbf{35}, 263--284 (2001).

\bibitem{ballester2022single}
M.~Ballester, H.~Wang, J.~Li, O.~Cossairt, and F.~Willomitzer, \enquote{Single-shot tof sensing with sub-mm precision using conventional cmos sensors,} {\protect\JournalTitle{arXiv preprint arXiv:2212.00928}}  (2022).

\bibitem{slogsnat2009non}
E.~Slogsnat and K.-H. Brenner, \enquote{Non-stereoscopic method for deflectometric measurement of reflecting surfaces,} {\protect\JournalTitle{DGaO-Proceedings (Online-Zeitschrift der Deutschen Gesellschaft f{\"u}r angewandte Optik e. V.), ISSN}} pp. 1614--8436 (2009).

\bibitem{huang2017zonal}
L.~Huang, J.~Xue, B.~Gao, C.~Zuo, and M.~Idir, \enquote{Zonal wavefront reconstruction in quadrilateral geometry for phase measuring deflectometry,} {\protect\JournalTitle{Applied optics}} \textbf{56}, 5139--5144 (2017).

\bibitem{wang2021vr}
J.~Wang, B.~Xu, T.~Wang, W.~J. Lee, M.~Walton, N.~Matsuda, O.~Cossairt, and F.~Willomitzer, \enquote{Vr eye-tracking using deflectometry,} in \emph{Computational Optical Sensing and Imaging,}  (Optica Publishing Group, 2021), pp. CF2E--3.

\bibitem{wang2023accurate}
J.~Wang, T.~Wang, B.~Xu, O.~Cossairt, and F.~Willomitzer, \enquote{Accurate and fast vr eye-tracking using deflectometric information,} in \emph{Computational Optical Sensing and Imaging,}  (Optica Publishing Group, 2023), p. CTh2B.5.

\bibitem{wang2022easy}
J.~Wang, B.~Xu, O.~Cossairt, and F.~Willomitzer, \enquote{Easy and flexible calibration approach for deflectometry-based vr eye-tracking systems,} in \emph{Computational Optical Sensing and Imaging,}  (Optica Publishing Group, 2022), pp. CTh5C--1.

\bibitem{ren2015iterative}
H.~Ren, F.~Gao, and X.~Jiang, \enquote{Iterative optimization calibration method for stereo deflectometry,} {\protect\JournalTitle{Optics express}} \textbf{23}, 22060--22068 (2015).

\bibitem{zhang2000flexible}
Z.~Zhang, \enquote{A flexible new technique for camera calibration,} {\protect\JournalTitle{IEEE Transactions on pattern analysis and machine intelligence}} \textbf{22}, 1330--1334 (2000).

\bibitem{angelopoulos2021event}
A.~N. Angelopoulos, J.~N. Martel, A.~P. Kohli, J.~Conradt, and G.~Wetzstein, \enquote{Event-based near-eye gaze tracking beyond 10,000 hz,} {\protect\JournalTitle{IEEE Transactions on Visualization and Computer Graphics}} \textbf{27}, 2577--2586 (2021).

\bibitem{ramanauskas2006calibration}
N.~Ramanauskas, \enquote{Calibration of video-oculographical eye-tracking system,} {\protect\JournalTitle{Elektronika Ir Elektrotechnika}} \textbf{72}, 65--68 (2006).

\bibitem{harezlak2014towards}
K.~Harezlak, P.~Kasprowski, and M.~Stasch, \enquote{Towards accurate eye tracker calibration--methods and procedures,} {\protect\JournalTitle{Procedia Computer Science}} \textbf{35}, 1073--1081 (2014).

\bibitem{nystrom2013influence}
M.~Nystr{\"o}m, R.~Andersson, K.~Holmqvist, and J.~Van De~Weijer, \enquote{The influence of calibration method and eye physiology on eyetracking data quality,} {\protect\JournalTitle{Behavior research methods}} \textbf{45}, 272--288 (2013).

\bibitem{corbett1994topography}
M.~C. Corbett, D.~P. O'Brart, D.~C. Saunders, and E.~S. Rosen, \enquote{The topography of the normal cornea,} {\protect\JournalTitle{European journal of Implant and Refractive Surgery}} \textbf{6}, 286--297 (1994).

\bibitem{atchison2004eye}
D.~A. Atchison, C.~E. Jones, K.~L. Schmid, N.~Pritchard, J.~M. Pope, W.~E. Strugnell, and R.~A. Riley, \enquote{Eye shape in emmetropia and myopia,} {\protect\JournalTitle{Investigative Ophthalmology \& Visual Science}} \textbf{45}, 3380--3386 (2004).

\bibitem{wang2023optimization}
T.~Wang, J.~Wang, O.~Cossairt, and F.~Willomitzer, \enquote{Optimization-based eye tracking using deflectometric information,} {\protect\JournalTitle{arXiv preprint arXiv:2303.04997}}  (2023).

\bibitem{cossairt2020low}
O.~Cossairt, F.~Willomitzer, C.-K. Yeh, and M.~Walton, \enquote{Low-budget 3d scanning and material estimation using pytorch3d,} in \emph{2020 54th Asilomar Conference on Signals, Systems, and Computers,}  (IEEE, 2020), pp. 1316--1317.

\end{thebibliography}

\end{document}